\documentclass[runningheads]{llncs}
\usepackage[T1]{fontenc}
\usepackage{graphicx}
\usepackage{amsmath}
\usepackage{amssymb}
\usepackage{booktabs}
\usepackage{float}
\usepackage{xcolor}
\usepackage{placeins}

\usepackage{algorithm}
\usepackage{algpseudocode}
\usepackage{graphicx}

\begin{document}
\title{Adversarially Guided Diffusion for LiDAR Range Image Synthesis}
%
%
\author{Stavros Bouras\inst{1}\orcidID{0009-0009-2948-1833} \and
Antonios Makris\inst{1}\orcidID{0000-0003-0514-4292} \and
Alexandros Gkillas\inst{2}\orcidID{0000-0001-5339-2018} \and
Aris S. Lalos\inst{2}\orcidID{0000-0003-0511-9302} \and
Konstantinos Tserpes\inst{1}\orcidID{0000-0001-5183-1443}}
\authorrunning{S. Bouras et al.}
%
\institute{School of Electrical and Computer Engineering, National Technical University of Athens, Greece \email{\{stavros\_bouras,antoniosmakris,tserpes\}@mail.ntua.gr} \and
Industrial Systems Institute, Athena Research Center, Patras Science Park, Greece
\email{\{gillas,lalos\}@athenarc.gr}
}
\maketitle              
\begin{abstract}

LiDAR semantic segmentation is a key perception task in autonomous driving, where false predictions can affect downstream planning and safety-critical decision-making. Although adversarial attacks, and specifically adversarial examples, have been widely studied for image classification and 3D point cloud segmentation, unrestricted adversarial examples remain largely unexplored in the space of 2D range images, which are projections of 3D point clouds. The proposed method is, to the best of our knowledge, the first diffusion-based unrestricted adversarial attack against 2D range-image segmentation, using adversarial guidance from a segmentation loss. By applying guidance directly during sampling, the method produces unrestricted adversarial examples that remain close to the learned LiDAR data manifold while inducing structured segmentation errors. Experiments on the SemanticKITTI dataset using RangeNet++ and CENet segmentation networks demonstrate that the attack provides adjustable degradation across guidance strengths and transfers across segmentation architectures. Compared with norm-bounded FGSM and SegPGD baselines, the proposed attack offers a distinct effectiveness–realism trade-off, achieving controllable white-box and transfer degradation while maintaining competitive distributional and visual realism.

\end{abstract}
\section{Introduction}

Deep Learning (DL) models are increasingly deployed in real-world, safety-critical perception tasks such as semantic segmentation \cite{guo2018review}, where reliable scene understanding is required under strict operational constraints \cite{zhang2019review}. In the automotive domain specifically, segmentation networks must process live data sourced from Light Detection and Ranging (LiDAR) sensors to understand the surrounding scene. LiDAR sensors provide accurate depth information in the form of 3D point clouds, making them particularly valuable for autonomous driving perception. Semantic segmentation can then be applied to these LiDAR points, or to LiDAR-derived representations, to assign semantic labels to relevant scene elements and support accurate, robust, and real-time scene understanding in complex driving environments \cite{feng2020deep}.

LiDAR semantic segmentation can be performed directly on the 3D point cloud \cite{hu2020randla} or on their 2D range image projection space \cite{milioto2019rangenet++}. 
This projection is widely adopted in the literature, offering a computationally efficient alternative to direct 3D point-cloud processing and enabling more scalable operations in 2D space for embedded automotive systems.
However, given their reliance on complex DL architectures, LiDAR semantic segmentation networks inherit the susceptibility of conventional deep neural networks to adversarial examples \cite{szegedy2013intriguing,goodfellow2014explaining}. These adversarial inputs are deliberately crafted perturbations designed to mislead the model while remaining imperceptible or structurally inconspicuous. For LiDAR semantic segmentation, such attacks may take the form of point perturbation, point injection, point removal, or physically realizable object-based manipulations \cite{zhu2021adversarial}. These attacks can corrupt the predicted segmentation map by producing large, spatially coherent misclassification regions. As a result, erroneous semantic outputs may propagate to downstream autonomous-driving modules, affecting obstacle avoidance, braking, lane keeping, and path planning \cite{cao2019adversarial}.

To address this challenge, recent work leverages generative models, particularly diffusion models, which produce realistic, high-fidelity samples and can be steered during sampling through guidance. This makes it possible to construct adversarial examples that fool the target model while remaining on the data manifold, rather than relying on small perturbations of a fixed input. So far, this approach has been demonstrated for RGB image classification and single-label classification \cite{dai2024advdiff,chen2023advdiffuser}, as well as for LiDAR segmentation in the 3D point-cloud domain \cite{wang2026transferable}.

To the best of our knowledge, this paper presents the first diffusion-based unrestricted adversarial attack on LiDAR semantic segmentation in the 2D range-image space. Our work steers a latent diffusion model using the gradient of a dense per-pixel segmentation loss, thus injecting  adversarial objectives directly into the reverse diffusion process. As a result, the generation is being guided towards realistic LiDAR range images that mislead the segmentation process while preserving the geometric structure of genuine scenes.

The contributions of our work can be summarized as follows:
\begin{itemize}
    \item We introduce the first diffusion-based adversarial attack on LiDAR range image semantic segmentation, guiding a latent diffusion model with the gradient of a segmentation loss during the reverse process. 
    \item We show that the attack offers controllable degradation across a range of attack strengths, while retaining strong transferability to black-box architectures.
    \item We show that the generated adversarial scenes can remain statistically close to real LiDAR data, demonstrating the unrestricted perturbations can preserve realism.
\end{itemize}

The remainder of the paper is organized as follows:
Section \ref{sec:preliminaries_and_relatedWork} presents the necessary preliminaries and reviews related work on adversarial diffusion generation and adversarial examples on semantic segmentation.
Section \ref{sec:proposed_attack} presents the proposed adversarially guided diffusion-based attack. Section \ref{sec:experiments} presents the experimental setup and evaluates the results.
Finally, Section \ref{sec:conclusions} concludes with key insights and outlines directions for future work.

\section{Preliminaries \& Related Work}
\label{sec:preliminaries_and_relatedWork}
\subsection{Adversarial Examples}

\textbf{Bounded Adversarial Examples.} Adversarial examples, which were first formalized for image classification \cite{szegedy2013intriguing,goodfellow2014explaining}, are a type of inference-time adversarial attacks that introduce small, imperceptible perturbations to the original clean samples with the goal of misleading a victim model into producing incorrect outputs. Popular algorithms for crafting adversarial examples include the Fast Gradient Sign Method (FGSM) \cite{goodfellow2014explaining}, which perturbs a given input $x$ by a single step in the direction that increases the training loss $\mathcal{L}$, so that $x' = x + \epsilon \cdot \mathrm{sign}(\nabla_x \mathcal{L}(x, y))$, with $\mathcal{L}(x, y)$ denoting the loss between the model's prediction on $x$ and the true label $y$, and $\nabla_x \mathcal{L}(x, y)$ its gradient with respect to the input. Building on this, Projected Gradient Descent (PGD) \cite{madry2017towards} applies this update iteratively. Optimization-based attacks have also been proposed for crafting adversarial examples~\cite{carlini2017towards}.

\textbf{Unrestricted Adversarial Examples.} The aforementioned attacks are restricted (i.e., bounded), since they apply a small perturbation to a fixed input $x$, subject to a closeness constraint (e.g., a fixed $\ell_p$ bound or a minimization of the perturbation norm). Unrestricted Adversarial Examples (UAEs) are a type of adversarial attack that moves beyond these constraints. Instead of perturbing a fixed input, they generate realistic new inputs from scratch (e.g., via generative models) that fool perception models while remaining on the benign data manifold \cite{song2018constructing}. Different generative models can be leveraged for this purpose, such as Generative Adversarial Networks (GANs) \cite{goodfellow2020generative} as presented in \cite{xiao2018generating}. However, since Diffusion Models (DMs) have been shown to outperform GANs on image generation \cite{dhariwal2021diffusion}, later implementations choose DMs as the generative model that drives the UAE generation \cite{dai2024advdiff,chen2023advdiffuser}.


\subsection{Diffusion Models}

\label{subsec:diffusion}
Diffusion models, established by Denoising Diffusion Probabilistic Models (DDPMs) \cite{ho2020denoising}, generate data by reversing a fixed forward noising process. The forward process is a fixed Markov chain that adds Gaussian noise according to a variance schedule $\beta_1,\dots,\beta_T$, with $q(\mathbf{x}_t \mid \mathbf{x}_{t-1}) := \mathcal{N}(\mathbf{x}_t; \sqrt{1-\beta_t}\,\mathbf{x}_{t-1}, \beta_t \mathbf{I})$, which allows $\mathbf{x}_t$ to be sampled directly from $\mathbf{x}_0$ as $\mathbf{x}_t = \sqrt{\bar\alpha_t}\,\mathbf{x}_0 + \sqrt{1-\bar\alpha_t}\,\boldsymbol{\epsilon}$, with $\boldsymbol{\epsilon} \sim \mathcal{N}(\mathbf{0}, \mathbf{I})$ and $\bar\alpha_t = \prod_{s=1}^{t}(1-\beta_s)$. The reverse process is a parameterized Markov chain starting from $\mathbf{x}_T \sim \mathcal{N}(\mathbf{0}, \mathbf{I})$, with $p_\theta(\mathbf{x}_{t-1} \mid \mathbf{x}_t) = \mathcal{N}(\mathbf{x}_{t-1}; \mu_\theta(\mathbf{x}_t, t), \Sigma_\theta(\mathbf{x}_t, t))$, implemented by training a network $\epsilon_\theta(\mathbf{x}_t, t)$ to predict the added noise. This noise-prediction objective is equivalent to score estimation, since $s_\theta(\mathbf{x}_t,t) = -\tfrac{1}{\sqrt{1-\bar\alpha_t}}\,\epsilon_\theta(\mathbf{x}_t,t) \approx \nabla_{\mathbf{x}_t}\log p_t(\mathbf{x}_t)$~\cite{ho2020denoising,song2020score}, where $p_t$ denotes the noisy data distribution at timestep $t$. This view enables steering the sampling trajectory, as will be detailed in Section~\ref{subsec:guidance}. To accelerate sampling, Denoising Diffusion Implicit Models (DDIMs)~\cite{song2020denoising} use the same training objective as DDPMs but replace the stochastic Markovian reverse process with a non-Markovian sampling procedure, enabling generation over a sparse subset of timesteps.

\subsection{Guidance in Diffusion Models}
\label{subsec:guidance}

DMs support both unconditional generation, where the model learns $p_{\mathrm{data}}(\mathbf{x})$, and conditional generation, where it learns $p_{\mathrm{data}}(\mathbf{x}\mid \mathbf{c})$ for a condition $\mathbf{c}$. Conditional generation uses the conditional score
$\nabla_{\mathbf{x}_t}\log p_t(\mathbf{x}_t\mid \mathbf{c})
=
\nabla_{\mathbf{x}_t}\log p_t(\mathbf{x}_t)
+
\nabla_{\mathbf{x}_t}\log p_t(\mathbf{c}\mid \mathbf{x}_t)$,
thereby guiding the reverse denoising process toward samples satisfying $\mathbf{c}$. The generally intractable guidance term $\nabla_{\mathbf{x}_t}\log p_t(\mathbf{c}\mid \mathbf{x}_t)$ is approximated in several ways: (i)~Classifier Guidance (CG), (ii)~Classifier-Free Guidance (CFG), and (iii)~Training-Free Guidance (TFG)~\cite{lai2025principles}. (i)~CG~\cite{dhariwal2021diffusion} trains an auxiliary time-conditional classifier $p_\psi(\mathbf{c} \mid \mathbf{x}_t, t)$, whose input gradient $\nabla_{\mathbf{x}_t} \log p_\psi(\mathbf{c} \mid \mathbf{x}_t, t) \approx \nabla_{\mathbf{x}_t} \log p_t(\mathbf{c} \mid \mathbf{x}_t)$ is used so that $\nabla_{\mathbf{x}_t} \log p_t(\mathbf{x}_t \mid \mathbf{c}, \omega) = \nabla_{\mathbf{x}_t} \log p_t(\mathbf{x}_t) + \omega \, \nabla_{\mathbf{x}_t} \log p_t(\mathbf{c} \mid \mathbf{x}_t)$, with $\omega$ the guidance scale. (ii)~CFG~\cite{ho2022classifier} achieves conditioning by modifying the gradient of the score, without relying on an explicit classifier, so that $\nabla_{\mathbf{x}_t} \log p_t(\mathbf{x}_t \mid \mathbf{c}, \omega) = \omega \, \nabla_{\mathbf{x}_t} \log p_t(\mathbf{x}_t \mid \mathbf{c}) + (1 - \omega) \, \nabla_{\mathbf{x}_t} \log p_t(\mathbf{x}_t)$, approximating those individual scores with neural networks. (iii)~TFG~\cite{ye2024tfg} leverages an external differentiable objective (e.g., a loss) whose gradient provides the guidance signal during the reverse denoising process. TFG can be injected in either the \textit{data space} or the \textit{noise space}~\cite{lai2025principles}: the former modifies the predicted clean sample $\hat{\mathbf{x}}_0$ so that $\hat{\mathbf{x}}_0(\mathbf{x}_t) + \eta_t \, \mathcal{G}_0$, with $\mathcal{G}_0 := -\nabla_{\mathbf{x}_0} \mathcal{L}(\mathbf{x}_0, \mathbf{c})$, while the latter modifies the score or noise prediction at timestep $t$ so that $\nabla_{\mathbf{x}_t} \log p_t(\mathbf{x}_t \mid \mathbf{c}) \approx \hat{\mathbf{s}}(\mathbf{x}_t) + \eta_t \, \mathcal{G}_t$, where $\eta_t$ controls the strength of the guidance correction. Data-space guidance avoids differentiating through the DM, unlike the noise-space correction.

\subsection{Related Work}
\textbf{Adversarial Diffusion Generation.} DMs can be leveraged to generate UAEs by incorporating adversarial objectives into the generation process. In the task of \textit{classification}, one line of work addresses this task by leveraging a DM as a generative prior while performing adversarial optimization over variables in its generation pipeline. This can be done either by optimizing inverted diffusion latents \cite{chen2024diffusion} or by applying an adversarial optimization attack (e.g., PGD) to the intermediate denoised samples at each step \cite{chen2023advdiffuser}. Another approach to generate UAEs with DMs is through the guidance mechanisms described in Section~\ref{subsec:guidance}. In this case, the reverse process is steered toward samples that remain realistic while inducing failure in a target model. The diffusion prior keeps the generated samples close to the learned data manifold, while the adversarial guidance term nudges them toward regions that degrade the target model's performance. As described in Section~\ref{subsec:guidance}, and following the classifier-guidance principle, a representative work presented in \cite{dai2024advdiff}, steered a class-conditional DM with the gradient of a target classifier in order to craft UAEs for image classification. In the context of \textit{3D point-cloud semantic segmentation}, the authors in \cite{wang2026transferable} followed the TFG paradigm and nudged the reverse process with a segmentation-loss gradient, bounding the generation pipeline with geometric constraints.

\textbf{Adversarial Examples on Semantic Segmentation.} Adversarial examples were extended from classification to semantic segmentation with Dense Adversary Generation (DAG)~\cite{xie2017adversarial}, which also demonstrated transferability in this setting. Subsequent work systematically evaluated segmentation robustness under $\ell_\infty$-bounded attacks~\cite{arnab2018robustness}, while stronger segmentation-specific methods such as MLAttack~\cite{gupta2019mlattack} and SegPGD~\cite{gu2022segpgd} were later proposed. Unrestricted adversarial examples have also been studied for 2D camera-image segmentation using a GAN-based approach~\cite{shen2019advspade}. However, diffusion-based unrestricted attacks for segmentation have so far been explored only in the 3D point-cloud domain~\cite{wang2026transferable}, leaving the 2D LiDAR range-image setting unaddressed. We fill this gap by developing diffusion-based UAEs that operate directly on LiDAR range images.

\section{Proposed Attack}
 \label{sec:proposed_attack}
 
\subsection{Overview}

Unlike prior unrestricted diffusion-based attacks, which have addressed image classification and, in the case of semantic segmentation, the 3D point-cloud domain, this work presents an unrestricted diffusion-based attack on semantic segmentation networks operating on 2D range images, i.e., the projection of LiDAR point clouds onto a 2D grid. Each LiDAR scan, originating from 3D scans, is represented as a 2D range image through spherical projection \cite{milioto2019rangenet++,xu2020squeezesegv3}. This projection assigns each 3D point to a pixel in a 2D grid based on the direction from which the LiDAR sensor observed it. The resulting image, of height $H$ and width $W$, consists of rows and columns that correspond to the sensor's vertical and horizontal viewing angles, and pixels that store the distance of the corresponding point from the sensor.

The goal of the proposed attack is to generate unrestricted adversarial examples (UAEs), namely adversarial 2D LiDAR range images targeting semantic segmentation networks. In semantic segmentation, a model receives a range image as input and predicts a class label for each pixel. Our attack aims to generate samples that remain realistic and consistent with the clean data distribution, while inducing erroneous pixel-level predictions and degrading the target model’s performance. In contrast to perturbation-based attacks, the generated samples are designed to remain on the clean data manifold.

\subsection{Attack Formulation}
 \label{subsec:formulation}

Our attack builds on the TFG setting described in Section~\ref{subsec:guidance}, where an external differentiable loss can be used to steer the reverse diffusion process through its gradient, without the need of retraining the DM conditionally or relying on a separate noise-aware model. In our case, this loss is a dense segmentation loss that measures the failure of a surrogate segmentation network on the generated range images. As a result, the guidance term, expressed as the gradient of the aforementioned loss, steers the generation process toward samples that induce incorrect pixel-level predictions, rather than toward samples satisfying a clean condition. 

The proposed attack leverages a pretrained conditional denoising network $\epsilon_\theta(\mathbf{z}_t, t, \mathbf{c})$ of a latent DM, where $\mathbf{z}_t$ is the noisy latent at timestep $t$, and $\mathbf{c}$ is the semantic conditioning map. The conditioning map represents a dense label map $\mathbf{y} \in \{0, \dots, K-1\}^{H \times W}$, where each pixel is assigned a semantic class. The set of valid pixels (i.e., pixels not assigned the unlabeled class $0$) is denoted as $\Omega_{\text{valid}} = \{(i,j) : y_{ij} \neq 0\}$. For the computation of the adversarial signal, a pretrained segmentation  network $g_\psi$ takes as input a range image and outputs per-pixel class logits. Since the diffusion model operates in latent space, it relies on an encoder-decoder setup to transform range images into latent representations and reconstruct them back into image space. Therefore, evaluating the segmentation loss requires decoding the latent into a range image. Since the diffusion model leverages a decoder with a non-differentiable quantization step, we use a differentiable decode that sidesteps quantization, allowing gradients to propagate back to the latent representation in order to compute the guidance gradient. The decoded output is subsequently passed through the segmentation network $g_\psi$, obtaining the predictions that will be used to compute the adversarial objective. This objective quantifies the extent to which the generated range image induces incorrect pixel-level predictions from the segmenter.Two modes can be realized: the \textit{untargeted} mode and the \textit{targeted}. In the former, the objective pushes predictions away from the conditioning labels $\mathbf{y}$, whereas in the latter, it pushes predictions toward a target label map $\tilde{\mathbf{y}}$. Both are realized as a dense per-pixel cross-entropy, computed over the valid pixels $\Omega_{\text{valid}}$ in the untargeted mode and over the source-class pixels $\Omega_{\text{src}}$ in the targeted mode. The two objectives are presented below:
\begin{equation}
\begin{aligned}
\mathcal{L}_{\text{adv}}^{\text{untargeted}}(\mathbf{z}) &= \frac{1}{|\Omega_{\text{valid}}|} \sum_{(i,j) \in \Omega_{\text{valid}}} \mathrm{CE}\big(g_\psi(\mathbf{z})_{ij},\, y_{ij}\big), \\[6pt]
\mathcal{L}_{\text{adv}}^{\text{targeted}}(\mathbf{z}) &= -\frac{1}{|\Omega_{\text{src}}|} \sum_{(i,j) \in \Omega_{\text{src}}} \mathrm{CE}\big(g_\psi(\mathbf{z})_{ij},\, \tilde{y}_{ij}\big),
\end{aligned}
\label{eq:objectives}
\end{equation}
where $\Omega_{\text{src}} = \{(i,j) \in \Omega_{\text{valid}} : y_{ij} = c_{\text{src}}\}$ represents the pixels whose ground-truth label is the source class $c_{\text{src}}$, and $\mathrm{CE}$ denotes the per-pixel cross-entropy.

The attack nudges the latent obtained after a deterministic DDIM update. The deterministic ($\eta=0$) sampler omits the per-step stochastic term of the reverse process, making the sampling trajectory fully determined by the initial latent $\mathbf{z}_T$ and the conditioning map $\mathbf{c}$. This also enables paired evaluation between clean counterparts $\mathbf{x}_{\mathrm{clean}}$, which can be generated from the same $\mathbf{z}_T$ and $\mathbf{c}$ with adversarial guidance disabled. The corresponding adversarial samples $\mathbf{x}_{\mathrm{adv}}$ will be generated in parallel with adversarial guidance enabled. As a result, differences between the two samples can be attributed to the guidance term rather than to  randomness in the sampling process. All the parameters for this paired evaluation are detailed in Section~\ref{subsec:exp_setup}.

Given the noise prediction $\hat{\boldsymbol{\epsilon}}_t = \epsilon_\theta(\mathbf{z}_t, t, \mathbf{c})$ and the clean latent estimate $\hat{\mathbf{z}}_0 = \frac{\mathbf{z}_t - \sqrt{1 - \bar\alpha_t}\,\hat{\boldsymbol{\epsilon}}_t}{\sqrt{\bar\alpha_t}}$, the deterministic DDIM step is obtained as
\begin{equation}
\mathbf{z}_{t-1}^{\,\text{ddim}} = \sqrt{\bar\alpha_{t-1}}\,\hat{\mathbf{z}}_0 + \sqrt{1 - \bar\alpha_{t-1}}\,\hat{\boldsymbol{\epsilon}}_t.
\label{eq:ddim_step}
\end{equation}

The adversarial guidance step is applied to the DDIM-updated latent only after a predefined portion of the reverse trajectory has elapsed. This choice leaves the early denoising steps unchanged, allowing the conditional diffusion model to first establish the global scene structure before the adversarial signal is introduced. This is particularly important in the TFG setting, where guidance is provided by an external surrogate segmentation network that is not trained to operate on noisy intermediate range images at every timestep \(t\). Activating the attack only after the sample has reached a sufficient level of realism and structure makes the surrogate loss more meaningful and stabilizes the guidance process.

The guidance, is then defined with respect to the current DDIM-updated  latent as follows:

\begin{equation}
\mathbf{g}_t = \nabla_{\mathbf{z}}\,\mathcal{L}_{\text{adv}}(\mathbf{z})\big|_{\mathbf{z} = \mathbf{z}_{t-1}^{\,\text{ddim}}}, \qquad
\mathbf{z}_{t-1} = \mathbf{z}_{t-1}^{\,\text{ddim}} + s \cdot \frac{\mathbf{g}_t}{\|\mathbf{g}_t\|_2}.
\label{eq:guidance}
\end{equation}
After the reverse sampling process is completed, the final clean and adversarial latents are mapped back to the 2D range-image space, providing $\mathbf{x}_{\text{clean}}$ and $\mathbf{x}_{\text{adv}}$, respectively. 

\begin{figure}[!t]
\centering
\includegraphics[width=\textwidth]{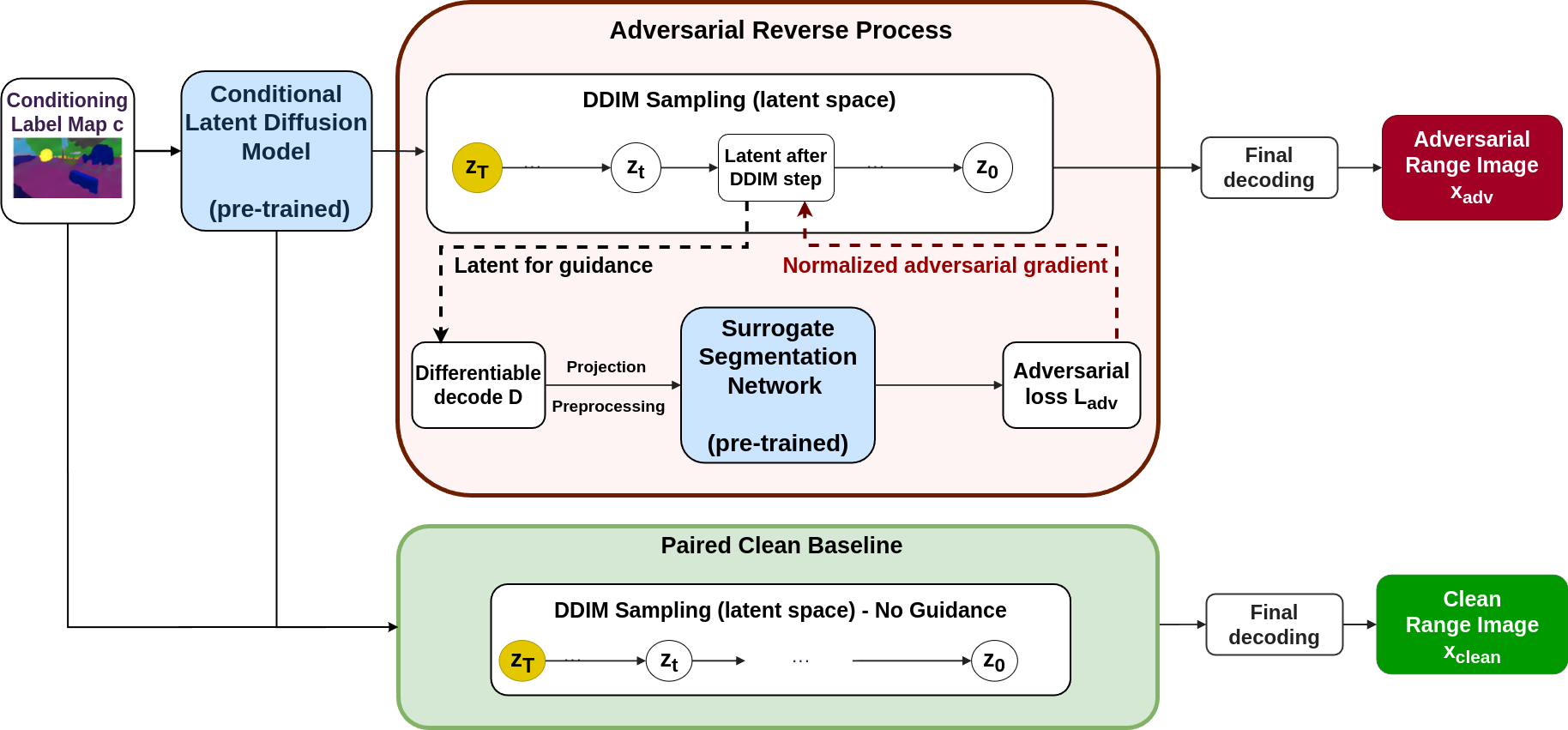}
\caption{Overview of the proposed attack.} 
\label{fig:overview}
\end{figure}

Figure \ref{fig:overview} illustrates the overall pipeline, with the upper branch depicting the adversarial reverse process and the lower branch the paired clean baseline, generated from the same initial noise. Finally, the complete conditional adversarial DDIM sampling process is summarized in Algorithm~\ref{alg:attack}.

\begin{algorithm}[!t]
\caption{Conditional Adversarial DDIM Sampling}
\label{alg:attack}
\begin{algorithmic}[1]
\Require pretrained conditional latent diffusion $\epsilon_\theta$ (frozen);
         frozen surrogate segmenter $g_\psi$;
         conditioning $\mathbf{c}$ with dense labels $\mathbf{y}$;
         DDIM steps $T$;
         guidance scale $s$;
         start fraction $\tau^\star$;
         terminal latent $\mathbf{z}_T \sim \mathcal{N}(\mathbf{0}, \mathbf{I})$
\Ensure adversarial range image $\mathbf{x}_{\text{adv}}$
\For{$t = T, T-1, \dots, 1$}
    \State $\hat{\boldsymbol{\epsilon}}_t \gets \epsilon_\theta(\mathbf{z}_t, t, \mathbf{c})$
        \Comment{noise prediction (no gradient)}
    \State $\hat{\mathbf{z}}_0 \gets \dfrac{\mathbf{z}_t - \sqrt{1 - \bar\alpha_t}\,\hat{\boldsymbol{\epsilon}}_t}{\sqrt{\bar\alpha_t}}$
        \Comment{predicted clean latent}
    \State $\mathbf{z}_{t-1}^{\,\text{ddim}} \gets \sqrt{\bar\alpha_{t-1}}\,\hat{\mathbf{z}}_0 + \sqrt{1 - \bar\alpha_{t-1}}\,\hat{\boldsymbol{\epsilon}}_t$
        \Comment{deterministic DDIM step, $\eta = 0$}
    \State $\tau \gets (T - t)/T$
    \If{$\tau \ge \tau^\star$}
        \State stop gradients into $\epsilon_\theta$; enable gradients on $\mathbf{z}_{t-1}^{\,\text{ddim}}$
        \State $\mathbf{g}_t \gets \nabla_{\mathbf{z}}\,\mathcal{L}_{\text{adv}}(\mathbf{z}) \big|_{\mathbf{z} = \mathbf{z}_{t-1}^{\,\text{ddim}}}$
        \State $\mathbf{z}_{t-1} \gets \mathbf{z}_{t-1}^{\,\text{ddim}} + s \cdot \dfrac{\mathbf{g}_t}{\|\mathbf{g}_t\|_2}$
            \Comment{$L_2$-normalized adversarial nudge}
    \Else
        \State $\mathbf{z}_{t-1} \gets \mathbf{z}_{t-1}^{\,\text{ddim}}$
            \Comment{no guidance ($\tau < \tau^\star$)}
    \EndIf
\EndFor
\State \Return $\mathbf{x}_{\text{adv}} = \mathcal{D}(\mathbf{z}_0)$
    
\end{algorithmic}
\end{algorithm}

\section{Experiments}
\label{sec:experiments}

\subsection{Experimental Setup}
\label{subsec:exp_setup}

\textbf{Dataset.} The evaluation was conducted on the SemanticKITTI dataset \cite{behley2019semantickitti}, a standard LiDAR semantic segmentation benchmark, which contains per-point class annotations across 20 classes, one of which was an ``unlabeled'' class, that was excluded from the evaluation of the experiments. The point clouds were projected to $64 \times 1024$ range images. Sequence 08 was the held-out sequence that was used for the validation across all experiments. It served both as the source for the conditioning label maps during the sampling process and as the \textit{real}-LiDAR reference distribution for realism comparison.

\textbf{Diffusion Model.} The proposed attack is developed on top of a pretrained latent diffusion model for LiDAR scene synthesis \cite{ran2024towards}. The diffusion model operates in a latent space, with a VQ-VAE encoder \cite{van2017neural} mapping range images to spatially-downsampled latent representations. Sampling was performed with deterministic DDIM ($\eta = 0$) for $T = 50$ denoising steps. For conditioning, the semantic-map-conditioned variant of the model, sem2lidar, was used. It was conditioned on a dense SemanticKITTI label map, producing $64 \times 1024$ range images.

\textbf{Segmentation networks.} Two off-the-shelf SemanticKITTI-pretrained segmentation networks were used and kept frozen: RangeNet++~\cite{milioto2019rangenet++} and CENet~\cite{cheng2022cenet}. The attack is evaluated by applying each segmenter to clean range images generated by the conditional DM and to their adversarial counterparts, and measuring the resulting degradation in segmentation performance. To evaluate both white-box and transfer settings, one network is used as the surrogate whose gradients guide sampling, while the other is used only for black-box evaluation. The networks are not fine-tuned on generated samples, avoiding circularity in the evaluation. Since both networks are pretrained on real scans, they yield lower absolute  \textit{mean Intersection over Union (mIoU)} scores when segmenting clean generated samples than when segmenting real scans. Since the scope of this work is to propose an attack that enables adversarial guidance in the sampling process of diffusion models operating on LiDAR data, we report attack effectiveness as the relative mIoU reduction from the clean generated baseline, which isolates the effect of the perturbation and is invariant to the absolute segmentation quality of the generator. We additionally report per-class IoU, where the absolute values for the dominant scene classes remain substantial and reveal how the degradation is distributed across the semantic categories.

\textbf{Generation and Evaluation.} Each surrogate network guides the generation of 1024 conditional samples. The evaluation examines both attack effectiveness and the realism of the generated adversarial samples. For the former, the relative white-box \textit{mIoU} reduction on the surrogate (WB drop \%) and on the transfer network is reported (Transfer drop \%), together with the per-class IoU, which shows how the degradation is distributed across the semantic classes. For the latter, realism is measured with a FID-based perceptual metric, the Fr\'echet Range Image Distance (FRID) \cite{heusel2017gans,ran2024towards}, computed over 1024 real scans from sequence 08. To avoid circularity, the FRID feature extractor is a segmentation network (RangeNet++) different from the surrogate that guided the sampling (CENet). In addition, metrics that measure the maximum
absolute per-pixel deviation between clean and adversarial samples
($L_\infty$), the mean squared per-pixel deviation over the sample (MSE) and the geometric distortion measured as the Chamfer Distance (CD) complement the realism evaluation.

\textbf{Proposed attack parameters.} The proposed attack exposes two adjustable parameters. The \textit{guidance scale} $s$ controls the magnitude of the adversarial nudge applied to the latent at each guided step. The \textit{start fraction} $\tau^\star$ controls when guidance begins along the sampling trajectory: a smaller $\tau^\star$ applies guidance over more sampling steps, while a larger $\tau^\star$ restricts it to the final steps.

 \textbf{Baselines.} For comparison, two norm-bounded attacks that operate on the decoded range image space, Fast Gradient Sign Method (FGSM) \cite{goodfellow2014explaining} and SegPGD \cite{gu2022segpgd}, are considered. FGSM is used as the canonical single-step attack, while  SegPGD is considered as a state-of-the-art iterative attack developed specifically for semantic segmentation.  SegPGD has been shown to outperform baseline adversarial attacks such as PGD and other segmentation specific attacks, including DAG and MLAttack. Since our contribution lies in the unrestricted regime, these two baselines are not direct competitors but serve to establish a norm-bounded reference against which the behaviour of UAEs can be contrasted. FGSM is applied as a single step, while SegPGD uses 20 iterations with step size $\alpha = 0.01$, each evaluated at three perturbation budgets ($\epsilon \in \{0.016, 0.063, 0.120\}$ and $\{0.004, 0.016, 0.120\}$, respectively).

\textbf{Attack modes.} The proposed attack supports two attack modes: \emph{untargeted}, which maximizes the segmentation loss against the ground-truth labels, degrading the performance of the segmentation network across the scene, and \emph{targeted}, which drives the prediction of a chosen source region toward a specified target class. For the baseline comparison, the untargeted mode is used, since the baseline attacks also operate in an untargeted manner. The targeted functionality of the proposed method is demonstrated in Figure~\ref{fig:targeted}.

\subsection{Results}

Table~\ref{tab:attack_comparison} reports the overall effectiveness of the attack.
The proposed attack shows that increasing the guidance scale $s$ consistently
degrades performance on both the white-box and transfer target models,
indicating that $s$ provides control over the attack strength. In addition, the
realism metrics show the trade-off between increasing the guidance scale and
moving away from the real-data distribution. However, the attack can still
achieve strong distributional similarity to real data, with FRID scores close to
the clean generated baseline for $s=3$ (statistically indistinguishable) and $s=6$, while achieving a relative
performance drop of up to $46.8\%$ in the white-box setting and up to $40.1\%$
in the transfer setting.

The CD values further indicate that the proposed guided diffusion attack produces larger structured geometric changes than the norm-bounded baselines. For example, CD increases from $0.52$ at $s=3$ to $1.02$ at $s=6$, while the FRID remains close to the clean generated baseline. This suggests that moderate guidance can introduce meaningful geometric changes without substantially degrading distributional realism.

The results also indicate that, compared with FGSM, our attack achieves a favorable effectiveness-realism trade-off at $s=10$, reaching a WB drop of 64.2\%, slightly higher than FGSM $\epsilon=0.120$, which reaches 63.2\%. However, our attack's FRID of 136.1 is significantly more realistic than 207.5, with a lower Chamfer distance as well, 1.38 compared with 2.39. This indicates that adversarial guidance during generation can achieve strong white-box degradation without the large distributional shift observed for high-budget FGSM.

In addition, compared to SegPGD, the strongest white-box perturbation baseline, a key finding appears in the transferability of our attack at comparable white-box strength and realism. SegPGD at $\epsilon=0.004$ achieves a slightly higher white-box drop than the proposed method at $s=6$, 53.3\% versus 46.8\%, and a similar FRID, 106.9 versus 108.6. However, its transfer drop is substantially lower, 21.7\% compared with 40.1\% for the proposed method. This suggests that there are settings in which the proposed unrestricted diffusion-based attack under adversarial guidance produces samples that generalize more effectively across segmentation architectures than low-budget post-hoc perturbations. It therefore provides  a distinct unrestricted alternative with stronger transfer behavior under comparable realism.

\begin{figure}[htb]
  \centering
  \includegraphics[width=\textwidth]{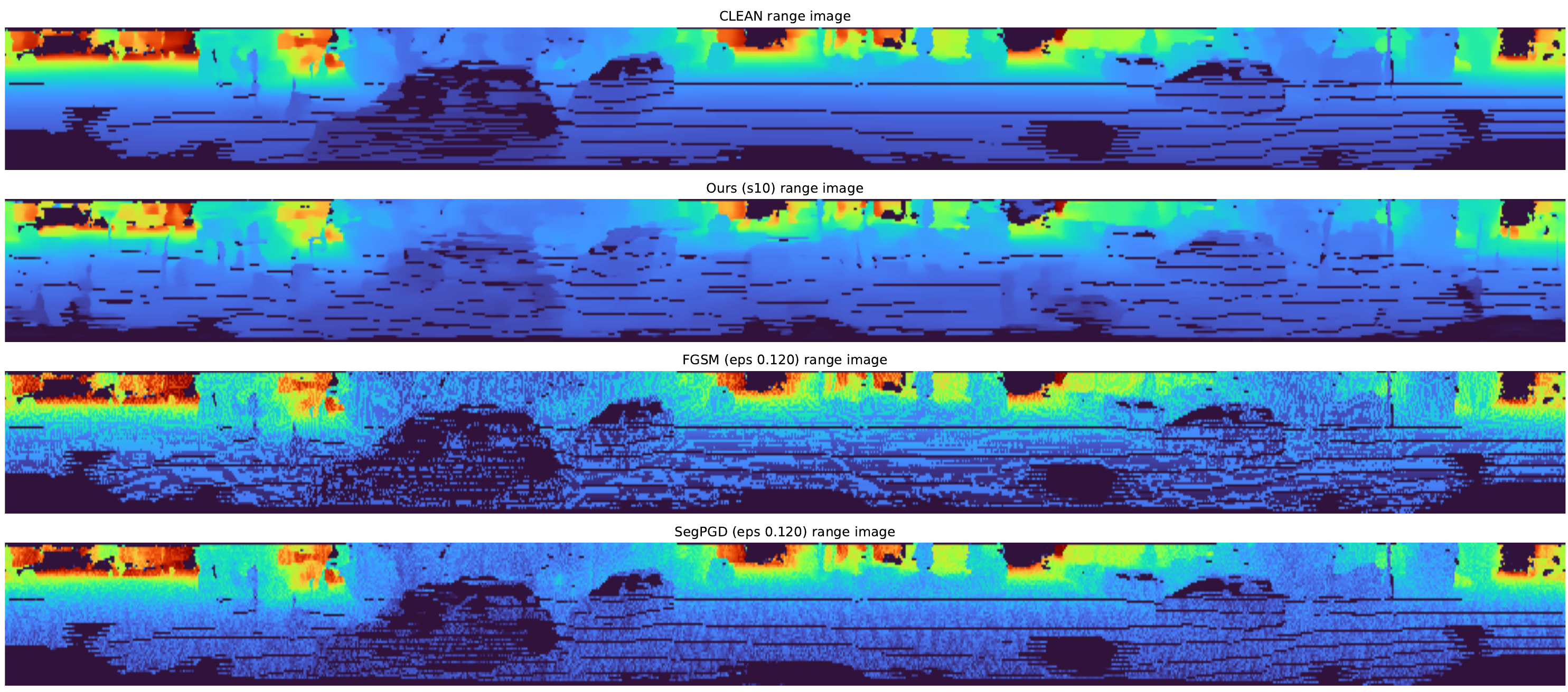}
  \caption{Range-image comparison for a single scene. From top to bottom: the clean generated range image, and three attacks at their high-strength settings (Ours $s=10$, FGSM $\epsilon=0.120$, SegPGD $\epsilon=0.120$). The norm-bounded baselines exhibit high-frequency noise across the range image, whereas the guided generation preserves smooth, scene-consistent structure.}
  \label{fig:comp_attacks}
\end{figure}

The interpretation of the results should be done under the scope that our proposed attack operates under a different regime than that of the two baselines. Ours is an unrestricted generative attack that modifies the reverse diffusion trajectory itself, while FGSM and SegPGD are post-hoc norm-bounded attacks applied to already-generated range images. Therefore the $L_\infty$ values should not be conceived as evidence of lower realism, but as an expected consequence of not enforcing a pointwise perturbation budget. In fact, prior diffusion-based unrestricted attack work~\cite{chen2024diffusion} showed that $L_\infty$ alone is an unsuitable indicator of human perception, and that in attacks constrained by a small perturbation budget, the perturbation can be easily perceived by the human eye as noticeable high-frequency noise. In contrast, the UAEs remain well-imperceptible. The claim made in \cite{chen2024diffusion} is also supported in our case, as shown in Figure~\ref{fig:comp_attacks}. In addition, in the same work, the authors noted that unrestricted attacks with larger $L_\infty$ values can achieve better transferability to black-box models, since they manipulate high-level semantics, which is also depicted in our results earlier. Finally, adversarial samples with lower $L_\infty$ have been shown to be more easily defended by purification techniques~\cite{nie2022diffusion,naseer2020self}.

Tables~\ref{tab:perclass_cenet} and~\ref{tab:perclass_rangenet} further demonstrate how the degradation is distributed across semantic classes, in both the CENet white-box setting (the surrogate) and the RangeNet++ transfer setting, even though the latter is not used for gradient computation, confirming that the adversarial effect is not limited to the surrogate model.

\begin{table}[htb]
\centering
\caption{Attack comparison on SemanticKITTI (CENet surrogate, RangeNet++ FRID extractor, $n{=}1024$). mIoU reported as relative drop (\%) from the clean generated baseline (CENet 9.19, RangeNet++ 18.91). FRID is measured against real scans. }
\label{tab:attack_comparison}
\setlength{\tabcolsep}{4pt}
\begin{tabular}{lcccccc}
\toprule
Attack & WB drop\,\% & Transfer drop\,\% & $L_\infty$ & MSE & CD & FRID \\
\midrule
clean & -- & -- & -- & -- & -- & 106.9 \\
\midrule
Ours ($s{=}3$)  & 25.7 & 18.7 & 1.93 & 0.0052 & 0.52 & 98.6 \\
Ours ($s{=}6$)  & 46.8 & 40.1 & 1.98 & 0.0108 & 1.02 & 108.6 \\
Ours ($s{=}10$) & 64.2 & 58.4 & 2.00 & 0.0160 & 1.38 & 136.1 \\
\midrule
FGSM $\epsilon{=}0.016$ & 54.3 & 45.9 & 0.037 & 0.0002 & 0.23 & 109.8 \\
FGSM $\epsilon{=}0.063$ & 61.3 & 65.7 & 0.067 & 0.0030 & 1.08 & 135.5 \\
FGSM $\epsilon{=}0.120$ & 63.2 & 77.5 & 0.121 & 0.0109 & 2.39 & 207.5 \\
\midrule
SegPGD $\epsilon{=}0.004$ & 53.3 & 21.7 & 0.028 & 0.00001 & 0.10 & 106.9 \\
SegPGD $\epsilon{=}0.016$ & 91.6 & 60.0 & 0.033 & 0.0001 & 0.18 & 107.8 \\
SegPGD $\epsilon{=}0.120$ & 98.0 & 69.9 & 0.121 & 0.0036 & 0.77 & 133.4 \\
\bottomrule
\end{tabular}
\end{table}

\begin{table}[!htb]
\centering
\caption{Per-class IoU under the CENet white-box attack (absolute IoU, clean (unattacked generated baseline) and adversarial, $n{=}1024$). $s\{3,6,10\}$: our attack at guidance scale; F$=$FGSM, S$=$SegPGD at the given $\epsilon$.}
\label{tab:perclass_cenet}
\setlength{\tabcolsep}{3pt}
\footnotesize
\begin{tabular}{lcccccccccc}
\toprule
Class & clean & $s3$ & $s6$ & $s10$ & F.016 & F.063 & F.120 & S.004 & S.016 & S.120 \\
\midrule
car          & 42.06 & 19.51 & 5.83 & 1.20 & 4.65 & 2.28 & 1.18 & 3.76 & 0.01 & 0.00 \\
road         & 65.73 & 57.85 & 48.23 & 35.62 & 53.60 & 33.82 & 26.60 & 50.68 & 3.22 & 0.18 \\
sidewalk     & 37.39 & 29.02 & 20.46 & 11.20 & 1.98 & 13.09 & 18.34 & 5.40 & 2.55 & 0.10 \\
building     & 13.38 & 12.42 & 11.25 & 9.73 & 11.59 & 12.21 & 12.16 & 12.83 & 6.72 & 2.23 \\
fence        & 1.68 & 1.36 & 0.84 & 0.65 & 0.75 & 1.11 & 1.02 & 0.96 & 0.33 & 0.23 \\
traffic-sign & 1.87 & 1.68 & 0.71 & 0.50 & 0.86 & 0.58 & 0.39 & 1.23 & 0.10 & 0.02 \\
\bottomrule
\end{tabular}
\end{table}

\begin{table}[!htb]
\centering
\caption{Per-class IoU under the RangeNet++ transfer evaluation (absolute IoU, clean and adversarial, $n{=}1024$). $s\{3,6,10\}$: our attack at guidance scale; F$=$FGSM, S$=$SegPGD at the given $\epsilon$.}
\label{tab:perclass_rangenet}
\setlength{\tabcolsep}{3pt}
\footnotesize
\begin{tabular}{lcccccccccc}
\toprule
Class & clean & $s3$ & $s6$ & $s10$ & F.016 & F.063 & F.120 & S.004 & S.016 & S.120 \\
\midrule
car        & 68.52 & 51.31 & 31.32 & 16.59 & 53.16 & 26.82 & 10.23 & 67.35 & 57.98 & 19.61 \\
road       & 66.18 & 48.39 & 24.99 & 9.04 & 10.51 & 4.49 & 3.24 & 30.72 & 13.52 & 3.92 \\
sidewalk   & 44.78 & 35.27 & 23.78 & 12.50 & 3.73 & 0.72 & 0.58 & 17.62 & 5.92 & 0.62 \\
building   & 43.59 & 39.79 & 35.02 & 29.08 & 29.59 & 16.35 & 9.87 & 44.01 & 35.43 & 12.22 \\
vegetation & 65.97 & 63.24 & 58.89 & 52.15 & 57.08 & 48.76 & 42.34 & 67.47 & 58.81 & 45.44 \\
terrain    & 20.10 & 16.24 & 12.70 & 11.20 & 10.76 & 8.82 & 6.42 & 12.16 & 11.77 & 10.58 \\
\bottomrule
\end{tabular}
\end{table}

\subsection{Complementary results}

\textbf{Surrogate independence.} To verify that the attack does not depend on the specific guiding network, a complementary experiment with RangeNet++ as the surrogate was conducted , evaluating realism with an independent CENet feature extractor. The attack reaches comparable white-box degradation (46.4\%) to the CENet-surrogate setting (46.8\%), and the resulting FRID (1.97) remains close to that of the clean samples (1.43), confirming that both effectiveness and realism are independent of the choice of surrogate. As this FRID is computed in CENet feature space, its scale differs from the RangeNet++-based FRID reported elsewhere and is only comparable within this experiment.

\textbf{Effect of the \textit{start fraction $\boldsymbol{\tau^\star}$}.} The start fraction $\tau^\star$ controls when guidance begins during sampling. Lower values of $\tau^\star$ apply guidance earlier and therefore guide a larger fraction of the trajectory, whereas higher values delay guidance and preserve more of the unguided generative process. For a fixed guidance scale of $s=6$, starting guidance earlier at $\tau^\star=0.25$ increases the attack strength compared with the default $\tau^\star=0.5$, raising the white-box drop from $46.8\%$ to $53.6\%$ and the transfer drop from $40.1\%$ to $51.2\%$. This increases FRID from $108.6$ to $123.2$, indicating reduced realism. Conversely, for $s=10$, delaying guidance to $\tau^\star=0.75$ reduces FRID from $136.1$ to $117.2$, but lowers the white-box and transfer drops from $64.2\%$ to $50.5\%$ and from $58.4\%$ to $36.0\%$, respectively. Overall, $\tau^\star$ controls the trade-off between attack effectiveness and sample realism.

\begin{figure}[!htb]
\centering
\includegraphics[width=\textwidth]{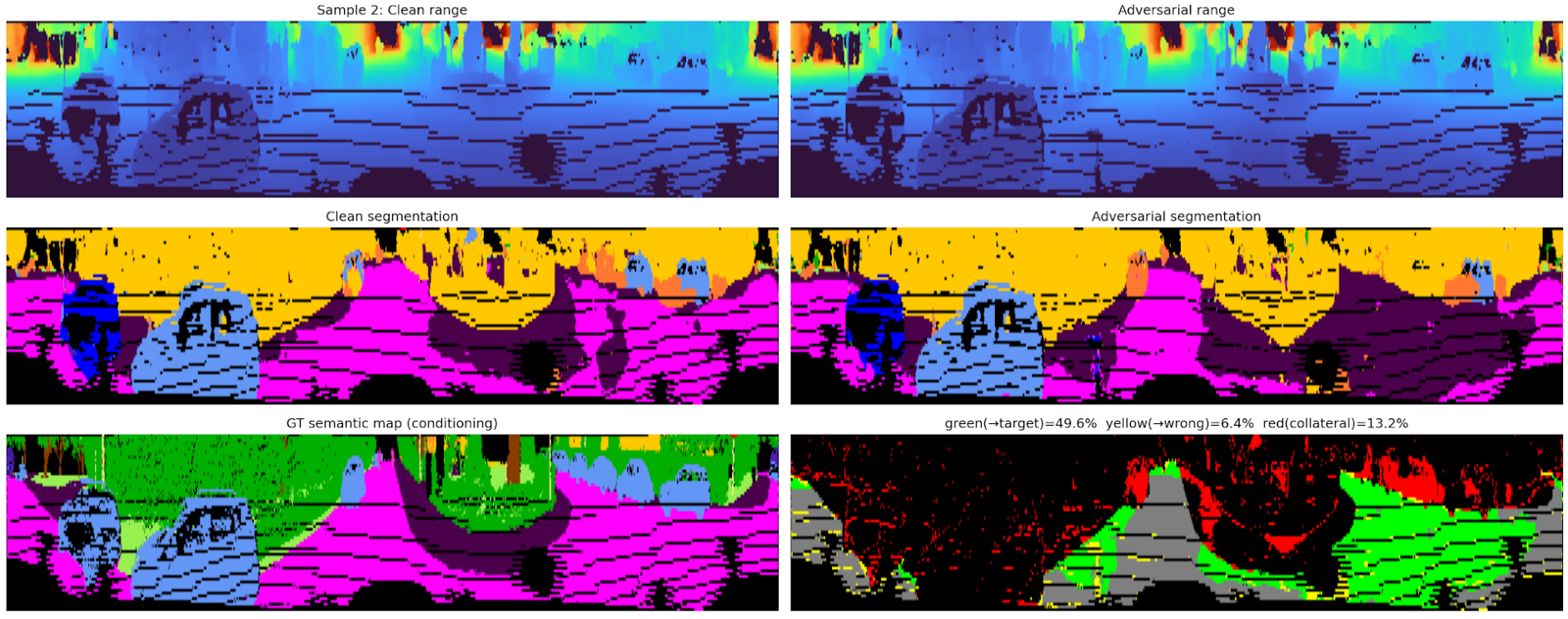}
\caption{Targeted attack example: Top row: clean / adversarial range images. Middle row: clean / adversarial segmentations. Bottom row: the conditioning semantic map and a per-pixel outcome map for the source (road) region. Outcome map: Green: road pixels reclassified as the target class sidewalk, Yellow:road pixels reassigned to other classes, Grey: road pixels that remain road, Red: collateral changes on non-source pixels. Of clean road predictions, $35.8\%$ are reclassified as sidewalk, $13.1\%$ as other classes, and $51.1\%$ remain road.}
\label{fig:targeted}
\end{figure}

\textbf{Targeted Attack functionality.} To showcase the functionality of the \textit{targeted} attack, Figure~\ref{fig:targeted} illustrates the targeted mode for an attack that aims to reclassify road pixels as sidewalk.



\FloatBarrier

\section{Conclusions}
\label{sec:conclusions}

This paper introduces, to the best of our knowledge, the first diffusion-based unrestricted adversarial attack for LiDAR range-image semantic segmentation. Unlike norm-bounded attacks that apply post-hoc perturbations to existing inputs, the proposed method steers the reverse trajectory of a conditional latent diffusion model using adversarial guidance from a dense segmentation loss. Experiments show controllable white-box and transfer degradation through the guidance parameters, while the generated range images remain realistic and close to the learned data distribution. Compared with FGSM and SegPGD, the method offers a complementary unrestricted adversarial regime in the 2D range-image space, producing smoother, scene-consistent samples, avoiding high-frequency artifacts, and showing stronger transferability in some settings.

Future work will explore the broader design space of adversarial guidance, including different injection points along the reverse sampling trajectory and alternative guidance-scaling strategies. A complementary direction is to train a noise-aware surrogate segmenter on noisy range images, which could provide more reliable gradients across the entire reverse trajectory.


%
%

\subsubsection{\ackname} This paper has received funding from the European Union’s Horizon Europe research and innovation actions under grant agreement
No 101168560 (CoEvolution). Views and opinions expressed are however those of the author(s) only and do not necessarily reflect those of the European Union or the Commission. Neither the European Union nor the granting authority can be held responsible for them.

\bibliographystyle{splncs04}
\bibliography{references}

@article{guo2018review,
  title={A review of semantic segmentation using deep neural networks},
  author={Guo, Yanming and Liu, Yu and Georgiou, Theodoros and Lew, Michael S},
  journal={International journal of multimedia information retrieval},
  volume={7},
  number={2},
  pages={87--93},
  year={2018},
  publisher={Springer}
}

@article{feng2020deep,
  title={Deep multi-modal object detection and semantic segmentation for autonomous driving: Datasets, methods, and challenges},
  author={Feng, Di and Haase-Sch{\"u}tz, Christian and Rosenbaum, Lars and Hertlein, Heinz and Glaeser, Claudius and Timm, Fabian and Wiesbeck, Werner and Dietmayer, Klaus},
  journal={IEEE Transactions on Intelligent Transportation Systems},
  volume={22},
  number={3},
  pages={1341--1360},
  year={2020},
  publisher={IEEE}
}

@article{zhang2019review,
  title={A review of deep learning-based semantic segmentation for point cloud},
  author={Zhang, Jiaying and Zhao, Xiaoli and Chen, Zheng and Lu, Zhejun},
  journal={IEEE access},
  volume={7},
  pages={179118--179133},
  year={2019},
  publisher={IEEE}
}

@article{szegedy2013intriguing,
  title={Intriguing properties of neural networks},
  author={Szegedy, Christian and Zaremba, Wojciech and Sutskever, Ilya and Bruna, Joan and Erhan, Dumitru and Goodfellow, Ian and Fergus, Rob},
  journal={arXiv preprint arXiv:1312.6199},
  year={2013}
}

@article{goodfellow2014explaining,
  title={Explaining and harnessing adversarial examples},
  author={Goodfellow, Ian J and Shlens, Jonathon and Szegedy, Christian},
  journal={arXiv preprint arXiv:1412.6572},
  year={2014}
}

@inproceedings{hu2020randla,
  title={Randla-net: Efficient semantic segmentation of large-scale point clouds},
  author={Hu, Qingyong and Yang, Bo and Xie, Linhai and Rosa, Stefano and Guo, Yulan and Wang, Zhihua and Trigoni, Niki and Markham, Andrew},
  booktitle={Proceedings of the IEEE/CVF conference on computer vision and pattern recognition},
  pages={11108--11117},
  year={2020}
}

@inproceedings{milioto2019rangenet++,
  title={Rangenet++: Fast and accurate lidar semantic segmentation},
  author={Milioto, Andres and Vizzo, Ignacio and Behley, Jens and Stachniss, Cyrill},
  booktitle={2019 IEEE/RSJ international conference on intelligent robots and systems (IROS)},
  pages={4213--4220},
  year={2019},
  organization={IEEE}
}

@inproceedings{xu2020squeezesegv3,
  title={Squeezesegv3: Spatially-adaptive convolution for efficient point-cloud segmentation},
  author={Xu, Chenfeng and Wu, Bichen and Wang, Zining and Zhan, Wei and Vajda, Peter and Keutzer, Kurt and Tomizuka, Masayoshi},
  booktitle={European Conference on Computer Vision},
  pages={1--19},
  year={2020},
  organization={Springer}
}

@inproceedings{dai2024advdiff,
  title={Advdiff: Generating unrestricted adversarial examples using diffusion models},
  author={Dai, Xuelong and Liang, Kaisheng and Xiao, Bin},
  booktitle={European Conference on Computer Vision},
  pages={93--109},
  year={2024},
  organization={Springer}
}

@inproceedings{chen2023advdiffuser,
  title={Advdiffuser: Natural adversarial example synthesis with diffusion models},
  author={Chen, Xinquan and Gao, Xitong and Zhao, Juanjuan and Ye, Kejiang and Xu, Cheng-Zhong},
  booktitle={Proceedings of the IEEE/CVF International Conference on Computer Vision},
  pages={4562--4572},
  year={2023}
}

@article{wang2026transferable,
  title={Transferable Adversarial Attacks on 3D Point Cloud Semantic Segmentation via Diffusion Models in Autonomous Driving},
  author={Wang, Yizhou and Wu, Libing and Zhang, Zhuangzhuang and Huo, Lijuan and Feng, Jiaqi and Wang, Jing and Jin, Jiong},
  journal={IEEE Transactions on Consumer Electronics},
  year={2026},
  publisher={IEEE}
}

@article{madry2017towards,
  title={Towards deep learning models resistant to adversarial attacks},
  author={Madry, Aleksander and Makelov, Aleksandar and Schmidt, Ludwig and Tsipras, Dimitris and Vladu, Adrian},
  journal={arXiv preprint arXiv:1706.06083},
  year={2017}
}

@inproceedings{carlini2017towards,
  title={Towards evaluating the robustness of neural networks},
  author={Carlini, Nicholas and Wagner, David},
  booktitle={2017 ieee symposium on security and privacy (sp)},
  pages={39--57},
  year={2017},
  organization={Ieee}
}

@article{song2018constructing,
  title={Constructing unrestricted adversarial examples with generative models},
  author={Song, Yang and Shu, Rui and Kushman, Nate and Ermon, Stefano},
  journal={Advances in neural information processing systems},
  volume={31},
  year={2018}
}

@article{goodfellow2020generative,
  title={Generative adversarial networks},
  author={Goodfellow, Ian and Pouget-Abadie, Jean and Mirza, Mehdi and Xu, Bing and Warde-Farley, David and Ozair, Sherjil and Courville, Aaron and Bengio, Yoshua},
  journal={Communications of the ACM},
  volume={63},
  number={11},
  pages={139--144},
  year={2020},
  publisher={ACM New York, NY, USA}
}

@article{dhariwal2021diffusion,
  title={Diffusion models beat gans on image synthesis},
  author={Dhariwal, Prafulla and Nichol, Alexander},
  journal={Advances in neural information processing systems},
  volume={34},
  pages={8780--8794},
  year={2021}
}

@article{xiao2018generating,
  title={Generating adversarial examples with adversarial networks},
  author={Xiao, Chaowei and Li, Bo and Zhu, Jun-Yan and He, Warren and Liu, Mingyan and Song, Dawn},
  journal={arXiv preprint arXiv:1801.02610},
  year={2018}
}

@article{ho2020denoising,
  title={Denoising diffusion probabilistic models},
  author={Ho, Jonathan and Jain, Ajay and Abbeel, Pieter},
  journal={Advances in neural information processing systems},
  volume={33},
  pages={6840--6851},
  year={2020}
}

@article{song2020denoising,
  title={Denoising diffusion implicit models},
  author={Song, Jiaming and Meng, Chenlin and Ermon, Stefano},
  journal={arXiv preprint arXiv:2010.02502},
  year={2020}
}

@article{song2020score,
  title={Score-based generative modeling through stochastic differential equations},
  author={Song, Yang and Sohl-Dickstein, Jascha and Kingma, Diederik P and Kumar, Abhishek and Ermon, Stefano and Poole, Ben},
  journal={arXiv preprint arXiv:2011.13456},
  year={2020}
}

@article{lai2025principles,
  title={The principles of diffusion models},
  author={Lai, Chieh-Hsin and Song, Yang and Kim, Dongjun and Mitsufuji, Yuki and Ermon, Stefano},
  journal={arXiv preprint arXiv:2510.21890},
  year={2025}
}

@article{ye2024tfg,
  title={Tfg: Unified training-free guidance for diffusion models},
  author={Ye, Haotian and Lin, Haowei and Han, Jiaqi and Xu, Minkai and Liu, Sheng and Liang, Yitao and Ma, Jianzhu and Zou, James and Ermon, Stefano},
  journal={Advances in Neural Information Processing Systems},
  volume={37},
  pages={22370--22417},
  year={2024}
}

@article{ho2022classifier,
  title={Classifier-free diffusion guidance},
  author={Ho, Jonathan and Salimans, Tim},
  journal={arXiv preprint arXiv:2207.12598},
  year={2022}
}

@article{chen2024diffusion,
  title={Diffusion models for imperceptible and transferable adversarial attack},
  author={Chen, Jianqi and Chen, Hao and Chen, Keyan and Zhang, Yilan and Zou, Zhengxia and Shi, Zhenwei},
  journal={IEEE Transactions on Pattern Analysis and Machine Intelligence},
  volume={47},
  number={2},
  pages={961--977},
  year={2024},
  publisher={IEEE}
}

@inproceedings{ran2024towards,
    title={Towards Realistic Scene Generation with LiDAR Diffusion Models},
    author={Ran, Haoxi and Guizilini, Vitor and Wang, Yue},
    booktitle={Proceedings of the IEEE/CVF Conference on Computer Vision and Pattern Recognition},
    year={2024}
}

@inproceedings{behley2019semantickitti,
  title={Semantickitti: A dataset for semantic scene understanding of lidar sequences},
  author={Behley, Jens and Garbade, Martin and Milioto, Andres and Quenzel, Jan and Behnke, Sven and Stachniss, Cyrill and Gall, Jurgen},
  booktitle={Proceedings of the IEEE/CVF international conference on computer vision},
  pages={9297--9307},
  year={2019}
}

@article{heusel2017gans,
  title={Gans trained by a two time-scale update rule converge to a local nash equilibrium},
  author={Heusel, Martin and Ramsauer, Hubert and Unterthiner, Thomas and Nessler, Bernhard and Hochreiter, Sepp},
  journal={Advances in neural information processing systems},
  volume={30},
  year={2017}
}

@inproceedings{cheng2022cenet,
  title={Cenet: Toward concise and efficient lidar semantic segmentation for autonomous driving},
  author={Cheng, Hui--Xian and Han, Xian--Feng and Xiao, Guo--Qiang},
  booktitle={2022 IEEE international conference on multimedia and expo (ICME)},
  pages={01--06},
  year={2022},
  organization={IEEE}
}

@inproceedings{gu2022segpgd,
  title={Segpgd: An effective and efficient adversarial attack for evaluating and boosting segmentation robustness},
  author={Gu, Jindong and Zhao, Hengshuang and Tresp, Volker and Torr, Philip HS},
  booktitle={European Conference on Computer Vision},
  pages={308--325},
  year={2022},
  organization={Springer}
}

@inproceedings{gupta2019mlattack,
  title={MLAttack: Fooling semantic segmentation networks by multi-layer attacks},
  author={Gupta, Puneet and Rahtu, Esa},
  booktitle={German conference on pattern recognition},
  pages={401--413},
  year={2019},
  organization={Springer}
}

@article{nie2022diffusion,
  title={Diffusion models for adversarial purification},
  author={Nie, Weili and Guo, Brandon and Huang, Yujia and Xiao, Chaowei and Vahdat, Arash and Anandkumar, Anima},
  journal={arXiv preprint arXiv:2205.07460},
  year={2022}
}

@inproceedings{naseer2020self,
  title={A self-supervised approach for adversarial robustness},
  author={Naseer, Muzammal and Khan, Salman and Hayat, Munawar and Khan, Fahad Shahbaz and Porikli, Fatih},
  booktitle={Proceedings of the IEEE/CVF conference on computer vision and pattern recognition},
  pages={262--271},
  year={2020}
}

@inproceedings{xie2017adversarial,
  title={Adversarial examples for semantic segmentation and object detection},
  author={Xie, Cihang and Wang, Jianyu and Zhang, Zhishuai and Zhou, Yuyin and Xie, Lingxi and Yuille, Alan},
  booktitle={Proceedings of the IEEE international conference on computer vision},
  pages={1369--1378},
  year={2017}
}

@inproceedings{arnab2018robustness,
  title={On the robustness of semantic segmentation models to adversarial attacks},
  author={Arnab, Anurag and Miksik, Ondrej and Torr, Philip HS},
  booktitle={Proceedings of the IEEE conference on computer vision and pattern recognition},
  pages={888--897},
  year={2018}
}

@article{shen2019advspade,
  title={Advspade: Realistic unrestricted attacks for semantic segmentation},
  author={Shen, Guangyu and Mao, Chengzhi and Yang, Junfeng and Ray, Baishakhi},
  journal={arXiv preprint arXiv:1910.02354},
  year={2019}
}

@article{van2017neural,
  title={Neural discrete representation learning},
  author={Van Den Oord, Aaron and Vinyals, Oriol and others},
  journal={Advances in neural information processing systems},
  volume={30},
  year={2017}
}

@inproceedings{zhu2021adversarial,
  title={Adversarial attacks against lidar semantic segmentation in autonomous driving},
  author={Zhu, Yi and Miao, Chenglin and Hajiaghajani, Foad and Huai, Mengdi and Su, Lu and Qiao, Chunming},
  booktitle={Proceedings of the 19th ACM conference on embedded networked sensor systems},
  pages={329--342},
  year={2021}
}

@inproceedings{cao2019adversarial,
  title={Adversarial sensor attack on lidar-based perception in autonomous driving},
  author={Cao, Yulong and Xiao, Chaowei and Cyr, Benjamin and Zhou, Yimeng and Park, Won and Rampazzi, Sara and Chen, Qi Alfred and Fu, Kevin and Mao, Z Morley},
  booktitle={Proceedings of the 2019 ACM SIGSAC conference on computer and communications security},
  pages={2267--2281},
  year={2019}
}

\end{document}